# Deep Convolutional Neural Network for Automated Detection of Mind Wandering using EEG Signals

Seyedroohollah Hosseini
*Department of Computer Science and Engineering*
*University of North Texas*
Denton, Texas, USA
sh0773@unt.edu

Xuan Guo
*Department of Computer Science and Engineering*
*University of North Texas*
Denton, Texas, USA
xuan.guo@unt.edu

*Abstract*—Mind wandering (MW) is a ubiquitous phenomenon which reflects a shift in attention from task-related to task-unrelated thoughts. There is a need for intelligent interfaces that can reorient attention when MW is detected due to its detrimental effects on performance and productivity. In this paper, we propose a deep learning model for MW detection using Electroencephalogram (EEG) signals. Specifically, we develop a channel-wise deep convolutional neural network (CNN) model to classify the features of focusing state and MW extracted from EEG signals. This is the first study that employs CNN to automatically detect MW using only EEG data. The experimental results on the collected dataset demonstrate promising performance with 91.78% accuracy, 92.84% sensitivity, and 90.73% specificity.

*Keywords—mind wandering, deep learning, convolutional neural network, electroencephalogram*

## I. INTRODUCTION

Most people have had the experience of a shift in their focus from an attention-demanding task at hand, such as reading books or driving, to task-unrelated concerns. For instance, they may be thinking of events which happened in the past, may happen in the future, or never happen at all. This stimulus-independent experience, which is a series of imaginative thought intermittently during sustained attention task, is called mind wandering (MW) [1]. Although some research linked MW to increased creativity [2], other studies indicated that MW can lead to errors and a decrease in performance in various tasks [3]. Moreover, there are evidences that support the correlation between MW and emotional disorders, such as neuroticism, alexithymia, and dissociation [4], [5]. Given the ubiquity and negative consequences of MW, it is beneficial to detect when MW occurs and then intervene to restore attention to the task at hand. As an initial step in this direction, this paper proposes an automated model to detect momentary occurrences of MW.

MW detection is a relatively unexplored field. A majority of research have correlated oculometric measures such as blink rate [6], pupil diameter and response, and eye gaze [7][8] to MW. Zerhouni et al. [9] provided an online measure of MW using fMRI sampling. Recently, there is a trend to study MW using EEG signals. Kawashima et al. [10] utilized the combination of EEG variables and non-linear regression modeling as an indicator of MW. Qin et al. [11] characterized MW with an increased gamma band activity in EEG signals. Son et al. [12] showed an increase in frontal EEG theta/beta ratio during MW. However, the collected cognitive results by these techniques are inconclusive, disputable and sometimes contradictory [13][14]. That is why finding a data-driven technique that can detect MW accurately, efficiently is vital.

Deep learning techniques have been vastly used to perform tasks such as making predictions and decisions. Recently, researchers have applied CNN, one type of deep learning models, on EEG signals. For instance, Schirrmeister et al. [15] studied a range of different CNN architectures for EEG decoding and visualization. Hajinoroozi et al. [16] designed a CNN to predict driver fatigue using 37 test participants' EEG signals. Acharya et al. [17] utilized the CNN to diagnose epileptic seizure using EEG signals. Since EEG signals are nonlinear and nonstationary in nature, we propose to employ CNN for the automatic identification of MW only using raw EEG signals. To the best of our knowledge, this is the first study that employs CNN for automatic detection of MW based on EEG data. The order of content in this paper is as follows: section 2 describes the dataset used in this study; section 3 explains the architecture of the proposed CNN, and training and testing processes; section 4 shows the performance based on experimental results; and section 5 concludes the paper.

## II. DATASET PREPARATION

### A. Subjects and procedure

EEG data used in this research was acquired by Grandchamp et al. at the University of Toulouse (Data is available on: https://sccn.ucsd.edu/publicly_available_EEG data.html) [13]. Datasets were collected from two participants, a female age 25 and a male age 31 both without mental or neurological disorder. Here we summarized the data collection procedure. More details are available in the original paper [13]. Participants sat in a dimly lit room in front of a computer screen. The task of the participants was to count backward each of their breath cycles (inhale/exhale) from 10 to 1 repeatedly. The mental state for the backward counting is referred as focusing state (FS). Participants had to press a mouse button whenever they realized they had lost track of their breath count (i.e. MW happens). They then filled a short questionnaire describing their mind wandering episode and pressed a button to resume the task. Each participant took ten 20-min sessions performing the above procedure. EEG data were captured using a 64 channel Biosemi Active Two system. The sampling rate was 1024 Hz. The data was recorded continuously through the whole sessions.

### B. Training data selection

There are two common strategies to define training data and target labels: trial-wise and cropped training strategy [15]. Trial-wise strategy considers the entire trial as one input sample with multiple labels, which indicate the corresponding events. Cropped-training strategy extracts time windows from a trial and assigns only one label for each time window. The assigned labels describe the events happened in that time window. We use the cropped training strategy in this work. We extracted MW episodes as positive samples and FS episodes as negative samples. Fig. 1 shows a heatmap of a 10-second interval of one EEG data session. This time window contains 5 seconds before and after the button was pressed. From Fig. 1, we can identify three distinct time blocks: the first time block from the beginning to the point when the participants realized MW, the second time block until the button was pressed, and the last time block that the participant was answering the questionnaire. Fig. 1 clearly provides some interesting patterns of mental states for FS and MW.

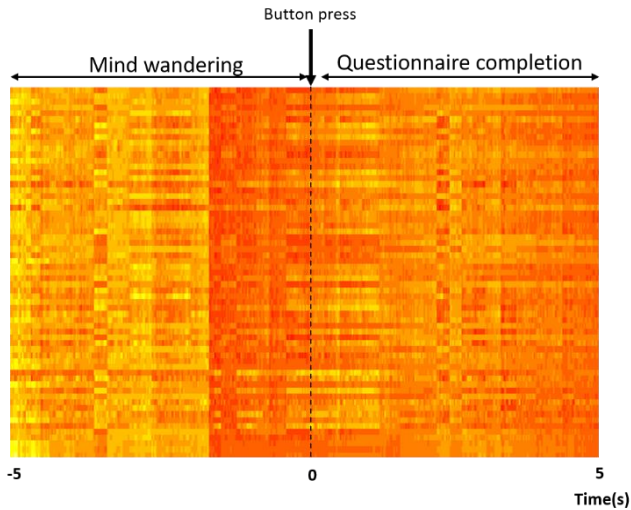

Fig. 1. Heatmap of 10-second EEG data. There is a change in EEG data before the button press.

We used MNE [18], an open-source python software, for training sample selection and data preparation (Source code is available on: https://github.com/Biocomputing-Research-Group/MW-Data-Selection). A total of 950 samples of 8 seconds were extracted from the original EEG data, 475 FS and 475 MW samples. We tried multiple time window lengths, and 8 seconds gave the best performance. More details are in section 4. The FS samples are the time intervals that the participants keep track of their breath counting. The FS samples were randomly chosen from original data after the breath counting started. Grandchamp et al [13] considered MW episodes as a 10-second interval prior to the button pressed. However, Fig. 1 shows that there is a clear change in the EEG signal before the button is pressed, which might be highly related to the mental activities responsible for realizing the occurrence of MW or hitting the button. Our observations show that this change lasts less than 2 seconds. Therefore, in order to have a clean sample with only MW signals, we considered an 8-second time window, starting 10 seconds before the button press. We also did a bandwidth filtering of 0.5 – 50 Hz, given the routine clinical EEG bandwidth is 0.5-50 Hz [19]. The extracted EEG samples were normalized by Z-score normalization with a zero mean and a standard deviation of 1.

## III. CNN MODEL

### A. Architechture

For the data sample representation, since EEG signals are the approximation of global voltage patterns from multiple sources of the brain, we consider not to combine these patterns, and use the whole set of electrodes. We also treat EEG as global modulation in time and space. In this study, we used the representation proposed by Schirrmeister et al. [15] which represents the input as a 2-D array where channels (electrodes) are rows and amplitude (voltage) across the time steps are columns.

The aim of the model is to automatically and explicitly extract features which are useful for EEG signal classification. Here, we propose a generic architecture that can reach a competitive accuracy. Our CNN has 12 layers including 4 convolution-max-pooling blocks, and three fully connected layers. Table 1 describes the details of the proposed CNN structure.

Table 1. Details of CNN structure used in this research

| Layers | Type | Kernel size | Stride | Output feature size |
|---|---|---|---|---|
| 1 | Convolution | $1 \times 11$ | 1 | $64 \times 8182 \times 20$ |
| 2 | Convolution | $64 \times 1$ | 1 | $1 \times 8182 \times 20$ |
| 3 | Max-pooling | $1 \times 2$ | 2 | $1 \times 4091 \times 20$ |
| 4 | Convolution | $1 \times 10$ | 1 | $1 \times 4082 \times 20$ |
| 5 | Max-pooling | $1 \times 4$ | 4 | $1 \times 1021 \times 20$ |
| 6 | Convolution | $1 \times 10$ | 1 | $1 \times 1012 \times 20$ |
| 7 | Max-pooling | $1 \times 4$ | 4 | $1 \times 253 \times 20$ |
| 8 | Convolution | $1 \times 11$ | 1 | $1 \times 243 \times 20$ |
| 9 | Max-pooling | $1 \times 3$ | 3 | $1 \times 81 \times 20$ |
| 10 | Fully-connected | - | - | 100 |
| 11 | Fully-connected | - | - | 50 |
| 12 | Fully-connected | - | - | 2 |

The first convolution layer receives input data. A kernel of size $1 \times 11$ performs a temporal convolution over time. It captures the correlation among signal in each channel. It outputs 20 sets of features. EEG signals produced by the brain have spatial correlation, meaning that all 64 channels capture the electrical activity of the brain where multiple brain regions might work together. Therefore, the second layer takes a spatial filtering with a kernel size of $64 \times 1$ to obtain the correlation among channels. The kernel size and stride were chosen by the trial and error. Our experiments showed that the similar kernel size and stride across these convolution layers can achieve better classification accuracy (data not shown in this paper). The rectified linear unit (RELU) function was used as activation function for each convolutional layer. During the training, we used a dropout layer with dropout rate 0.2 after the last max-pooling layer. The used loss function is cross entropy function. Backpropagation was performed to calculate gradients (partial derivatives) and update the parameters by Adam optimizer [20].

Classification block consists of three fully-connected layers. The first fully-connected layer gets the feature map output by the forth feature extraction block. This layer outputs 100 features. The second fully—connected layer generates 50 features. Finally, the third fully-connected layer produces 2 features that will be fed into Softmax function to generate the probability of each class label. Fig. 2 depicts the graphical representation of proposed CNN structure for MW detection.

### B. Training and testing

Ten-fold cross validation [21] was used to test the generalized performance of our CNN in classifying MW and FS. We randomly divided the whole data to ten equal portions (folds). Eight out of ten folds were used to train the model, one of the last two folds was used for validation, and the last one was utilized for testing the model. We repeated this procedure ten times such that all folds were used to test the model once. Each repetition of ten-fold cross validation includes 100 epochs, meaning that the number of iterations through the training set is 100 for backpropagation.

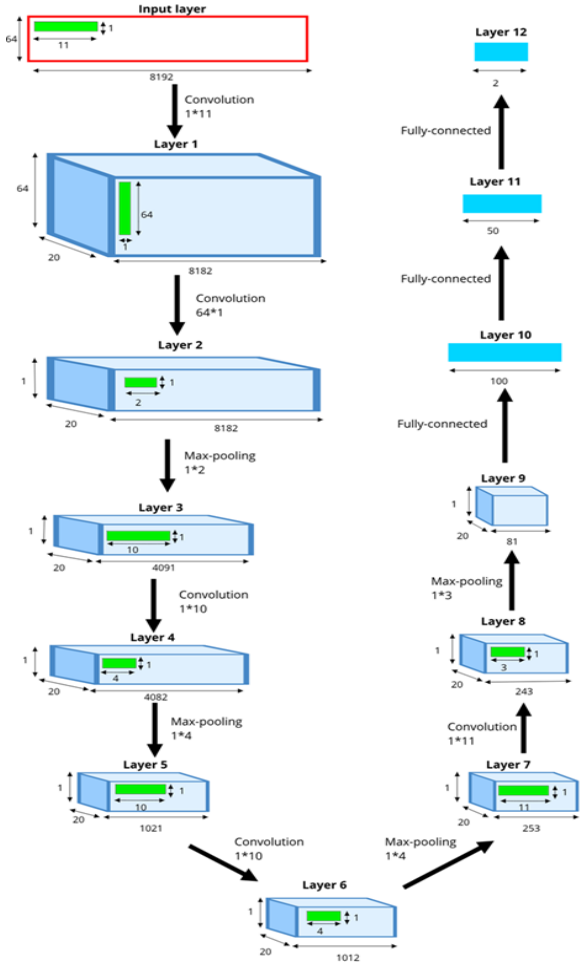

Fig. 2. The deep CNN architecture for MW detection

## IV. EXPERIMENTS AND RESULTS

The proposed CNN model was implemented on a workstation with Intel Xeon 2.6 GHz (Silver 4112), 32 GB random access memory (RAM), and 8 GB DDR5 graphic card (NVidia Quadro P4000) with 1792 CUDA cores using Python 3.6 and PyTorch machine learning library (Source code is available on: https://github.com/Biocomputing-Research-Group/MW-Detection).

Seven metrics were used to measure the performance: true positive, true negative, false positive, false negative, accuracy, sensitivity, and specificity. Positive means MW is identified by model, and negative means the model rejects the happening of MW. True positive is the total number of times the model correctly identifies MW. True negative counts the total number of times the model correctly rejects MW. False positive is the number of times the model incorrectly identifies the occurrence of MW. False negative is the total number of times the model incorrectly rejects the happening of MW. Accuracy quantifies the ability of model for the task of MW classification by averaging the truly classified samples over the total. Sensitivity measures the proportion of true positives. Specificity measures the proportion of true negatives. Table 2 presents the confusion matrix of the model results for the MW classification task.

Table 2. Overall classification result across 10 folds

| True positive | True negative | False positive | False negative | Accuracy | Sensitivity | Specificity |
|---|---|---|---|---|---|---|
| 441 | 431 | 34 | 44 | 91.78% | 92.84% | 90.73% |

### A. Performance comparison of varied time window lengths

In order to decide the best time interval of a data sample, we tried three different lengths: 2, 5, and 8 seconds. So, we obtained three trainning datasets. The same CNN model as in section 3 were used with max-pooling layers size modified. More specifically, all these CNN models have 4 max-pooling layers. The model for 8-second data samples has the first max-pooling size set to $1\times 2$ with stride 2, the second and the third layers set to $1\times 4$ with stride 4, and the forth layer set to $1\times 3$ with stride 3. The CNN model for 5-second has the first and the second max-pooling layers set to size $1\times 2$ with stride 2, the third layer set to $1\times 3$ with stride 3, and the forth layer set to $1\times 4$ with stride 4. The model for 2-second data samples has the first and second max-pooling layers set to size $1\times 2$ with stride 2, the third layer set to $1\times 3$ with stride 3, and the final max-pooling layer set to $1\times 2$ with stride 2.

The classification results are shown in Table 3. The results show that 8-second yielded the best classification rate. The 8-second gave the classification accuracy of 91.78%, while the classification rate for 5-second and 2-second are 86.63% and 78.52% respectively. However, the gain from changing 5-second to 8-second is marginal compared to the one from changing 2-second to 5-second.

Table 3. Classification accuracy for different sample lengths

| Sample length (second) | 2 | 5 | 8 |
|---|---|---|---|
| Classification accuracy (%) | 78.52 | 86.63 | 91.78 |

Fig. 3 shows the change of classification accuracy over the number of itrations for these datasets. Because the original paper [13] defined MW as 10-second time windows prior to the button press, and we excluded the 2-second time windows before the button press from our positive data samples due to the change in EEG heatmap (see Fig.1), the extracted 8-second time windows are the longest time windows that can be used as MW samples in this work.

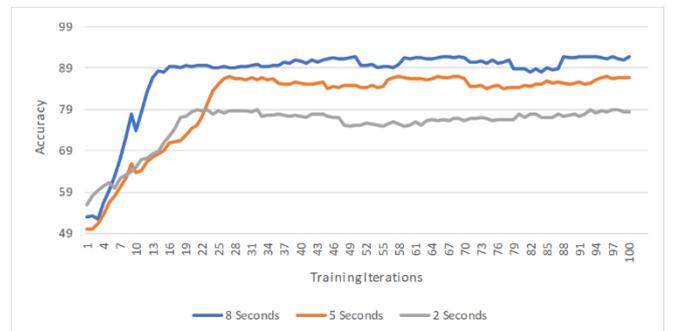

Fig. 3. Classification accuracy of the selected sample length

## B. Performance comparison with individual variation

There was an interest to figure out if different subjects might produce some patterns of MW specific to each subject. In other words, we want to test if there is any considerable feature-wise difference between our male and female participants regarding the MW detection. To do so, we employed the proposed CNN model in three classification runs. In the first run, we trained the CNN only using the EEG data captured from the male participant and tested the model using the data belongs to the female participant. In the second run, we did the training and testing in an opposite way that use the data from the female participant to train the model and test it using the data from the male participant. In the third run, we selected training and test samples randomly from the whole dataset including both male and female. To maintain the same sizes for training and testing, in these three runs, we divided the whole dataset to three parts, the first part containing 380 out of the 475 data samples (40% of whole dataset) to train the CNN, the second part containing 95 out of the 950 (10% of the whole dataset) for validation, and the third part including 475 data samples used to test the model.

The classification results are shown in Table 4. The model yielded the classification accuracies of 67.63% and 65.26% for the first two runs, respectively, and 81.84% classification accuracy for the third run. The performance lapse between the first two runs and the last run supports that there may be some unique features of MW or FS to individuals. Further study on these features related to MW at individual level are necessary which is beyond the scope of this work.

Table 4. Classification result for the experiment with gender factor

| Experiment run | Accuracy |
| --- | --- |
| First run | 67.63% |
| Second run | 65.26% |
| Third run | 81.84% |

## V. CONCLUSION

In this work, the deep CNN model is employed for the first time to detect MW using only EEG data. One advantage of this model is that it does not need to have an explicit step to extract features, which is integrated in the classification process. The model architecture was chosen by the trial and error. We considered three time window lengths and found 8-second give the best performance. The experimental results showed that our model can achieve the classification accuracy up to 91.78%. The significant drop of classification rates between each participant and the mixed data demonstrated that there might be some individual specific features of MW. For obtaining a good generalization performance, a diversity of data is required. The initial success of our model warrants our future study to apply the same deep learning approach on new datasets for automated MW detection.